\documentclass[10pt,twocolumn,letterpaper]{article}

\usepackage{iccv}              
\usepackage{multirow}
\usepackage{xcolor} 
\usepackage{colortbl} 
\usepackage{float}

\definecolor{iccvblue}{rgb}{0.21,0.49,0.74}
\usepackage[pagebackref,breaklinks,colorlinks,allcolors=iccvblue]{hyperref}

\def\paperID{2089} 
\def\confName{ICCV}
\def\confYear{2025}

\title{SRefiner: Soft-Braid Attention for Multi-Agent Trajectory Refinement}
\author{
Liwen Xiao$^{1}$\hspace{0.1in}  Zhiyu Pan$^{1}$\hspace{0.1in} Zhicheng Wang$^{1}$\hspace{0.1in}  Zhiguo Cao$^{1}$\hspace{0.1in} Wei Li$^{2}$\thanks{Corresponding author.} \\
$^1$School of AIA, Huazhong University of Science and Technology\\
$^2$S-Lab, Nanyang Technological University\hspace{0.3in}\\
{\tt\small \{liwenxiao\}@hust.edu.cn}
}

\begin{document}
\maketitle
\begin{abstract}
Accurate prediction of multi-agent future trajectories is crucial for autonomous driving systems to make safe and efficient decisions. Trajectory refinement has emerged as a key strategy to enhance prediction accuracy. However, existing refinement methods often overlook the topological relationships between trajectories, which are vital for improving prediction precision. Inspired by braid theory, we propose a novel trajectory refinement approach, Soft-Braid Refiner (SRefiner), guided by the soft-braid topological structure of trajectories using Soft-Braid Attention. Soft-Braid Attention captures spatio-temporal topological relationships between trajectories by considering both spatial proximity and vehicle motion states at ``soft intersection points". Additionally, we extend this approach to model interactions between trajectories and lanes, further improving the prediction accuracy. SRefiner is a multi-iteration, multi-agent framework that iteratively refines trajectories, incorporating topological information to enhance interactions within traffic scenarios. SRefiner achieves significant performance improvements over four baseline methods across two datasets, establishing a new state-of-the-art in trajectory refinement. Code is here https://github.com/Liwen-Xiao/SRefiner.
\end{abstract}    
\section{Introduction}

\begin{figure}[tp]
    \centering
    \includegraphics[width=\linewidth]{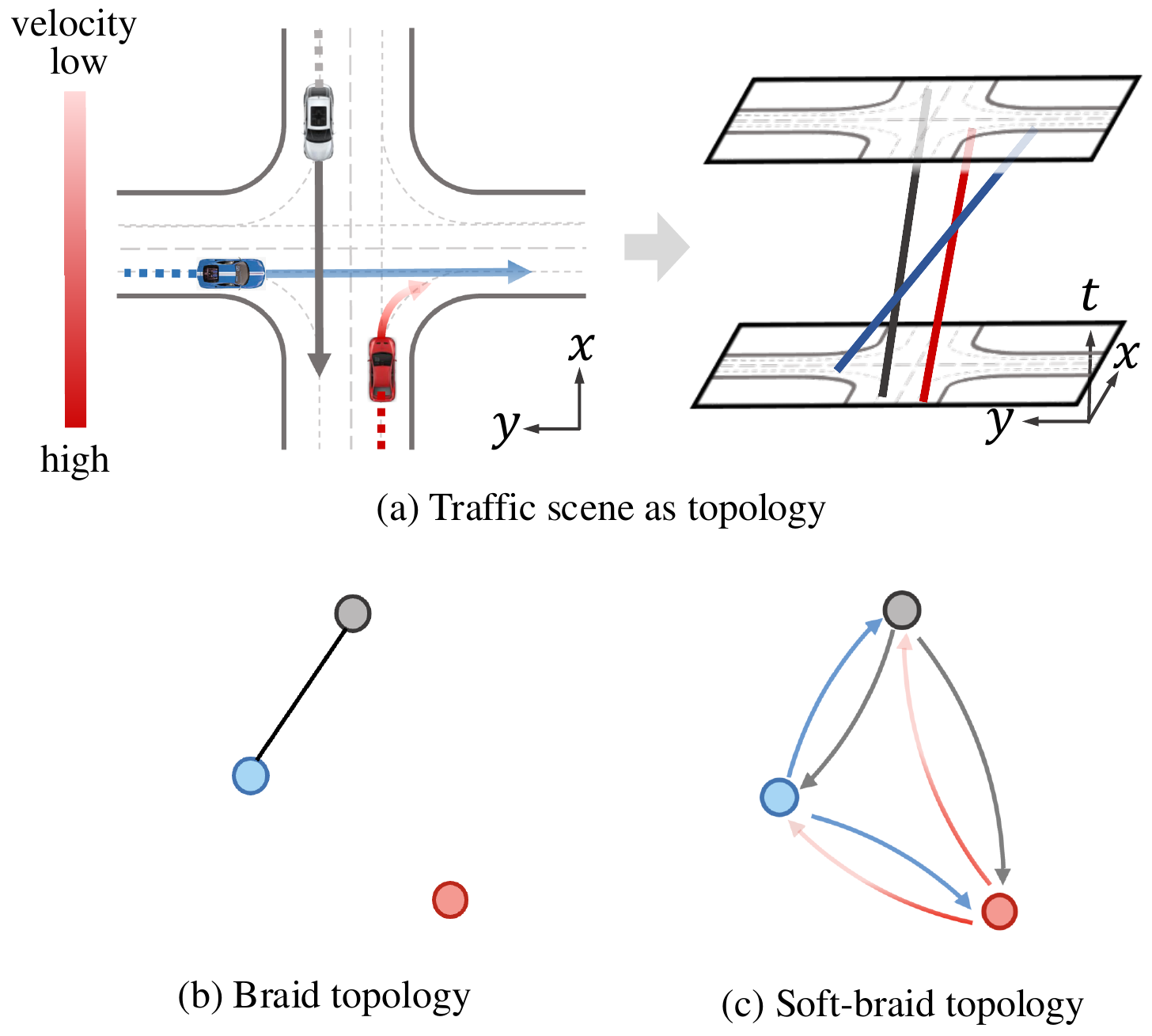}
    \caption{(a) Traffic scene with multi-agent future interaction. (b) Braid topology~\cite{braid1947theory} detects crossing situations, but fails to account for non-crossing interactions (e.g., red car decelerating for blue car in (a)). (c) We design soft-braid topology that establishes logical connections among all future trajectories, explicitly modeling spatio-temporal dynamics to guide refinement.}
    \label{fig:fig1}
\end{figure}

Predicting the future trajectories of surrounding traffic participants is essential for autonomous driving systems to make safe and efficient decisions in dynamic traffic environments~\cite{biff, densetnt, gameformer}. Consequently, accurate multi-agent trajectory prediction is crucial for enhancing the safety of autonomous vehicles. Recent studies have focused on improving the accuracy and rationality of multi-agent trajectory prediction from various perspectives, including scene representation~\cite{adapt, scenetransformer, tnt} and multi-agent behavior modeling~\cite{hivt, lanegcn, demo}.

To further improve prediction accuracy, trajectory refinement strategies have been widely adopted. These methods typically take an initially predicted future trajectory as input and produce a more accurate and reasonable output. For instance, QCNet~\cite{qcnet} encodes coarse predicted trajectories as anchor queries and combines them with scene context to predict trajectory offsets. R-Pred~\cite{rpred} utilizes local attention mechanisms to refine trajectories based on contextual information from neighboring vehicles.  SmartRefine~\cite{smartrefine} introduces a scene-adaptive refinement strategy that adjusts the number of refinement iterations and the range of local attention based on the scene's characteristics. MTR++~\cite{mtr++} leverages initial intent points to guide information exchange between traffic participants, enabling them to understand each other’s driving intentions. However, these trajectory refinement methods primarily model the interaction between vehicles and the environment through implicit relationship learning, without considering the explicit topological structure of vehicle trajectories, which has been proven to be effecient in robotics~\cite{hu2000optimal, wang2022coordination, mavrogiannis2024abstracting}, such as planning for tethered drones~\cite{cao2023neptune, cao2023path} and grasping~\cite{grasp}. Our goal is to model the explicit spatio-temporal topological representation of trajectories to guide the refinement process, specifically, with the help of braid theory~\cite{braid1947theory}.

Braid theory~\cite{braid1947theory} studies structures composed of multiple intertwined strands, known as braids, focusing on analyzing their topological relationships through crossing patterns like over-crossing and under-crossing. A direct application of braid theory~\cite{betop}, treats vehicle trajectories as strands in the braid and assumes that trajectories that cross each other are likely to influence one another. Therefore, during information fusion, only the interaction of information between crossing trajectories is enabled, thereby avoiding interference from irrelevant agents. as shown in Figure~\ref{fig:fig1}(b). We term it braid attention.

However, such direct application of braid topology to trajectory prediction tasks has three critical limitations‌:
(a) Neglecting some important interaction scenarios: For example, in Figure~\ref{fig:fig1}(a), although the trajectories of the red and blue vehicles do not intersect, the red vehicle slows down because it is waiting for the blue vehicle to pass. Their behaviors are logically correlated.
(b) Ignoring temporal dynamics: Braid attention focuses exclusively on spatial topological relationships between trajectories, overlooking their dynamic interactions over time.
(c) Limited expressiveness: Braid attention can only answer the question of \emph{whether trajectories interact} but fails to capture the details of \emph{how they interact}.

\begin{figure*}[thbp]
    \centering
    \includegraphics[width=\linewidth]{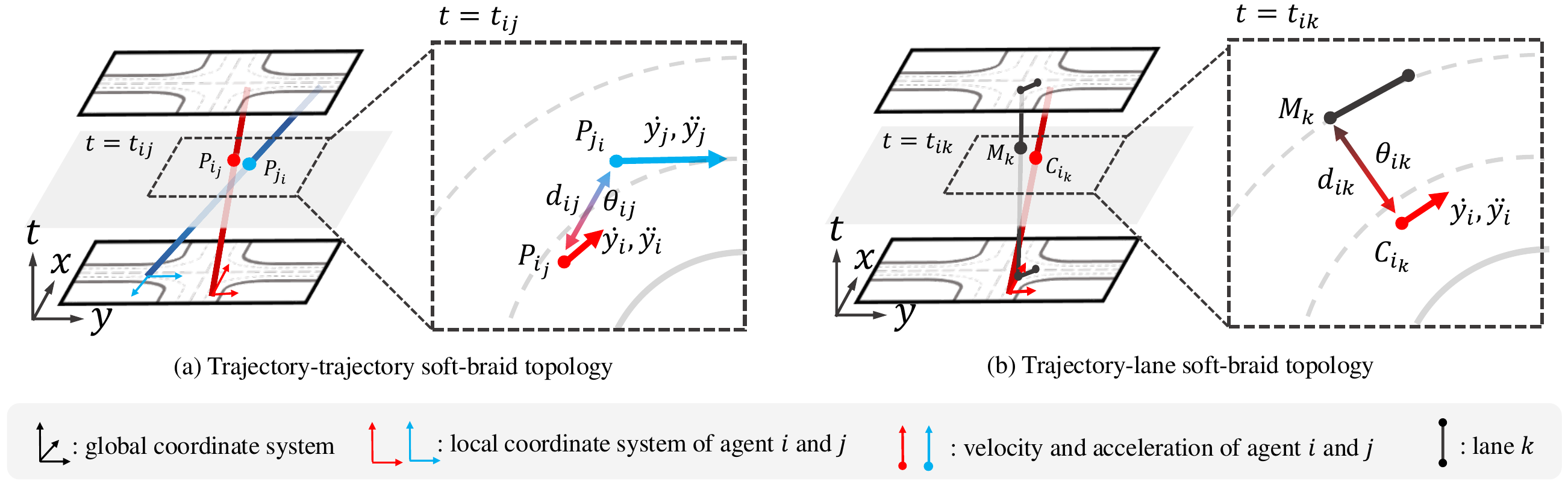}
    \caption{
    (a) 
    The two points that are closest to each other at the same moment on trajectory $ i $ and $ j $ are defined as the soft intersection points $ P_{i_j} $ and $ P_{j_i}$.
    The soft-braid topology between trajectories \(i\) and \(j\) is characterized by the velocity and acceleration of agents \(i\) and \(j\) at the soft intersection points, as well as the distance and angle of the line connecting the two points. The soft-braid topology of trajectory \(j\) relative to \(i\), is represented by these features expressed in the local coordinate systems of agents \(i\), and vice versa.
    (b) 
    The position where agent \(i\) and lane \(k\) are closest is defined as their soft intersection point $ C_{i_k}$. The soft-braid topology of lane \(k\) relative to trajectory \(i\) includes the velocity and acceleration of agent \(i\) at the soft intersection point, as well as the distance and angle of the line connecting the soft intersection point to lane \(k\), all expressed in the local coordinate system of agent \(i\).}
    \label{fig:softbraidtopology}
\end{figure*}

To overcome these limitations, we propose \textbf{Soft-Braid Attention}, which introduces the concept of ``soft intersection points". In our model, soft intersection points refer to the two points on the two trajectories that are closest to each other at the same moment. We consider not only the relative spatial relationship of these soft intersection points but also encode the vehicles’ motion states at these points. We term it as Trajectory-Trajectory Soft-Braid Topology. Furthermore, we extend soft-braid topology to interactions between trajectories and lanes, as shown in Figure~\ref{fig:softbraidtopology}. Specifically, we define the waypoint closest to a lane as the soft intersection point between the trajectory and the lane. The relative positional relationship and vehicle motion state at this point are encoded as Trajectory-Lane Soft-Braid Topology. These topology informations are integrated into subsequent interactions, benefiting trajectory refinement.

Based on this, we propose  \textbf{Soft-Braid Refiner (SRefiner)}, a multi-iteration multi-agent trajectory refinement framework. 
Through iterative interactions between trajectories and lanes, as well as between trajectories themselves, SRefiner ultimately delivers more accurate trajectory predictions. We validate the effectiveness of SRefiner on two datasets and four baseline methods. Experimental results demonstrate that SRefiner achieves significant accuracy improvements and establishes a new state-of-the-art (SOTA) performance in trajectory refinement.
Our contributions are summarized as follows:
\begin{itemize}
    \item We propose soft-braid attention, which explicitly models the spatio-temporal topological relationships between trajectories to guide their information fusion. This method is also capable of accommodating the modeling of interactions between trajectories and lanes.
    \item We propose Soft-Braid Refiner (SRefiner), an advanced multi-agent trajectory refinement framework that jointly refines the trajectories of all vehicles in a scene, enabling more precise scene prediction.
    \item We validate the effectiveness of SRefiner across multiple datasets and baselines, achieving SOTA performance in trajectory refinement tasks.
\end{itemize}

\section{Related Works}

\subsection{Multi-agent Motion Forecasting}
Multi-agent motion prediction can be divided into three categories: marginal prediction, conditional prediction, and joint prediction. Marginal prediction predicts the future trajectory of each agent independently, without considering the interaction between their future trajectories. These methods primarily focus on aspects such as scene representation~\cite{multipath, multipath++, tpcn, vectornet, covernet}, scene feature modeling~\cite{lanegcn, lanercnn, hivt, simpl, demo}, and scene normalization~\cite{scenetransformer, adapt, tnt}. Conditional prediction using the future states of another agent as an input. For instance, CBP~\cite{cbp} employs the ground truth future motion of the query agent and models the behavioral change of a target agent. M2I~\cite{m2i} learns to predict agent relationships by classifying them as influencer-reactor pairs and generates the reactor's trajectory based on the estimated influencer's trajectory. Joint prediction integrates the future features of multiple agents to simultaneously predict their future trajectories. FIMP~\cite{fimp} fuses multi-agent future feature based on their similarity. GANet~\cite{ganet} forecasts agents' possible future endpoint regions and aggregates the interaction features between the map and agents in these regions. BeTop~\cite{betop} enables information fusion among intersecting trajectories and disables it among non-intersecting ones. FutureNet-LOF~\cite{futurenet} predicts one future trajectory fragment at a time and uses it as the basis for predicting the next fragment, thereby modeling future interactions. 
However, existing motion forecasting networks still predict unreasonable trajectories, such as those involving collisions or leaving the drivable area. Consequently, recent works~\cite{dcms, qcnet, rpred, smartrefine, mtr++} have explored trajectory refinement methods. These methods take the rough future trajectories predicted by the baseline as input and output more reasonable future trajectories. Our SRefiner is a trajectory refinement method designed to address these issues.


\subsection{Trajectory Refinement}
Trajectory refinement methods aim to achieve more accurate and reasonable agent trajectories by conducting finely interactive modeling between the future trajectories proposed in the first stage and the scene. 
DCMS~\cite{dcms} achieves more spatially consistent refined trajectories by refining the proposed trajectory with added noise. QCNet~\cite{qcnet} re-encodes the proposed trajectory as an anchor query and then fuses the information again with the HD map and the historical states of agents in the scene. R-Pred~\cite{rpred} selects neighboring lanes and agents of the proposed trajectory through local attention for information interaction. MTR~\cite{mtr} also uses local attention and refines the trajectory through multiple iterations. Building on MTR, MTR++~\cite{mtr++} introduces mutually guided intention querying, enabling future trajectories of vehicles to interact with and influence each other's behavior. SmartRefine~\cite{smartrefine} introduces a scene-adaptive refinement strategy that dynamically adjusts the number of refinement iterations and the range of local attention to conduct fine-grained information interaction with the HD map. 
While existing trajectory refinement methods demonstrate promising results through implicit feature interactions, their reliance on latent space modeling fundamentally limits their capability to capture structured traffic priors. Such implicit interactions often obscure the precise contributions of different factors to the refined trajectories. In contrast, our method is motivated by the need for more effective and interpretable trajectory refinement. We aim to utilize 2D topological explicit features with strong expressiveness and interpretability to guide the refinement process.

\section{Method}

\subsection{Problem Formulation}


Given the historical trajectory coordinates $X \in \mathbb{R}^{N \times T_- \times 2} $ of $N$ agents in a scene and the high-definition (HD) map lanes $L$, a motion forecasting baseline can be expressed as 
\begin{equation}
    Y_0 = \mathcal{F}(X, L),
\end{equation}
where $\mathcal{F}(\cdot)$ represents the motion forecasting model, and $Y_0 \in \mathbb{R}^{K \times N \times T_+ \times 2}$ denotes the predicted future trajectory with $K$ modes. Here, $ T_- $ and $ T_+$ correspond to the historical and future horizons, respectively. We focus on the trajectory refinement stage, which aims to generate more accurate multi-agent joint trajectories by carefully modeling the interactions between the predicted future trajectory and its interactions with the HD map lanes:
\begin{equation}
    Y = Y_0 + \mathcal{R}(Y_0, L),
\end{equation}
where $\mathcal{R}(\cdot)$ is the refinement model. For the HD map $L$, we employ a vectorized representation. Each vector in $L$ encapsulates the coordinates and semantic details of a lane centerline. We denote the number of lane centerlines as $M$.  

\subsection{Soft Braid Topology}

\noindent\textbf{Braid topology.} In braid theory~\cite{braid1947theory}, a braid is represented as a tuple \(f = (f_1,\ldots,f_n)\) where each functions \(f_i: I \to \mathbb{R}^2\times I\) for \(i\in\mathcal{N}\) corresponds to a strand.
These founctions, defined on the interval \(I = [0,1]\), are embedded in the Cartesian space \((x,y,t)\) and increase monotonically along the \(t\)-axis. The topology relationship $ B $ is utilized to describe the intertwining relationships among the strands. 

\noindent\textbf{Modeling trajectories with braid topology.} In motion forecasting, each predicted future trajectory can be analogized to a strand in 2D space that evolves over time, denoted as $ y_i: t \to \mathbb{R}^2\times t $  for $1 \le i \le N$. The collection of these trajectories, $y = (y_1,\ldots, y_N)$, defined over the interval $t = [0,T_+]$, represent all the trajectories in a future scenario.
The topological intertwining relationships among these trajectories are captured by $ B = \{ \sigma_{i \leftarrow j}\}_{n \times n}, 1\le i,j \le n$. 
Specifically, if strand $f_i$ intersects with $f_j$ and $f_i$ is below $f_j$, then $\sigma_{i \leftarrow j}=1$. 
This relationship can be expressed as:
\begin{equation}
\sigma_{i \leftarrow j} = \begin{cases} 
1 & \exists\, 0<t_i<t_j<T_+:\ \big\| y_i(t_i) - y_j(t_j) \big\| < \varepsilon \\
0 & \text{otherwise},
\end{cases}
\end{equation}
where $\varepsilon$ is a threshold, typically representing the width of the vehicles.
$\sigma_{i \leftarrow j} = 1$ indicates that within the interval $ t \in [0, T_+] $, trajectories $y_i$ and $y_j$ intersect, with $y_i$ reaching the intersection first. This scenario is interpreted as agent $j$ yielding agent $i$. In braid attention, if $\sigma_{i \leftarrow j} = 1$ or $\sigma_{j \leftarrow i} = 1$, the future trajectories of agent $i$ and $j$ are considered to influence each other, the information interaction between $y_i$ and $y_j$ is enabled; otherwise, not. 

\noindent\textbf{Soft-braid topology.}
Directly applying braid topology to represent vehicle trajectories in autonomous driving presents several limitations. For instance, braid topology can only capture interactive scenarios where trajectories overlap, thereby neglecting important situations such as the one illustrated in Figure~\ref{fig:fig1}(a). Furthremore, braid topology merely indicates \emph{whether} trajectories interact without detailing \emph{how} they interact. Additionally, it falls to model dynamic interactions over time. To overcome these limitations, we propose a more suitable approach for autonomous driving trajectory prediction, soft-braid topology. This method effectively captures the intricate dynamic interactions between trajectories in the temporal domain.

Formally, we define the points on \(y_i\) and \(y_j\) that are closest to each other at the same time as soft intersection points \(P_{i_j}\) and \(P_{j_i}\), respectively. The time at which this minimum distance occurs is denoted as \(t_{ij}\):
\begin{equation}
    t_{ij} = \mathop{\arg\min}\limits_{t} ||y_i(t) - y_j(t) ||,
\end{equation}
\begin{equation}
    P_{i_j} = y_i(t_{ij}), \quad P_{j_i} = y_j(t_{ij}).
\end{equation}
We represent the spatial relationship between \( P_{i_j} \) and \( P_{j_i} \) using the distance \( d_{ij} \) between them and the orientation \( \theta_{ij} \) of the line connecting them:
\begin{equation}
    d_{ij} = ||P_{j_i} - P_{i_j}||, \quad \theta_{ij} = \arctan (P_{j_i} - P_{i_j}).
\end{equation}
We characterize the Soft-Braid topology between \(y_i\) and \(y_j\) using the motion states of agent \(i\) and agent \(j\) at the soft intersection points,
along with the positional relationship between these points:
\begin{equation}
    \tilde{\sigma}_{i \leftarrow j} = [ \dot{y}^{(i)}_i(t_{ij}), \dot{y}^{(i)}_j(t_{ij}), \ddot{y}^{(i)}_i(t_{ij}), \ddot{y}^{(i)}_j(t_{ij}), d_{ij}, \theta_{ij}^{(i)}],
\end{equation}
\begin{equation}
    \tilde{\sigma}_{j \leftarrow i} = [ \dot{y}^{(j)}_j(t_{ij}), \dot{y}^{(j)}_i(t_{ij}), \ddot{y}^{(j)}_j(t_{ij}), \ddot{y}^{(j)}_i(t_{ij}), d_{ij}, -\theta_{ij}^{(j)}],
\end{equation}
where $ \tilde{\sigma}_{i \leftarrow j} $ represents the soft-braid topology of \(y_j\) relative to \(y_i\). It includes the velocities $ \dot{y} $ and accelerations $ \ddot{y} $ of agent \(i\) and agent \(j\) in the local coordinate system of agent \(i\), along with the spatial positional relationship between the two soft intersection points. The same applies to $ \tilde{\sigma}_{j \leftarrow i} $.
Here, $(i)$ denotes the local coordinate system of agent \(i\). Its origin \(O_i\in\mathbb{R}^2\) is the endpoint of the its historical trajectory in the global coordinate system, and \(\theta_i\in\mathbb{R}\) represents its orientation in the global coordinate system at the end of the historical trajectory.
The transformation of a trajectory \(y\) from the global coordinate system to the local coordinate system of agent \(i\) is formulated as follows:
\begin{equation}
    y^{(i)} = (y - O_i) \begin{bmatrix} \cos \theta_i & -\sin \theta_i \\ \sin \theta_i & \cos \theta_i \end{bmatrix}.
\end{equation}
The same process applies to (j). $ \widetilde{B} = \{ \tilde{\sigma}_{i \leftarrow j} \}_{N \times N}, 1 \le i,j \le N$ represents the soft-braid topology among all $N$ trajectories in a future scenario.

\subsection{Soft-Braid Attention}
\label{Soft Braid Attention}

\begin{figure*}[tp]
    \centering
    \includegraphics[width=\linewidth]{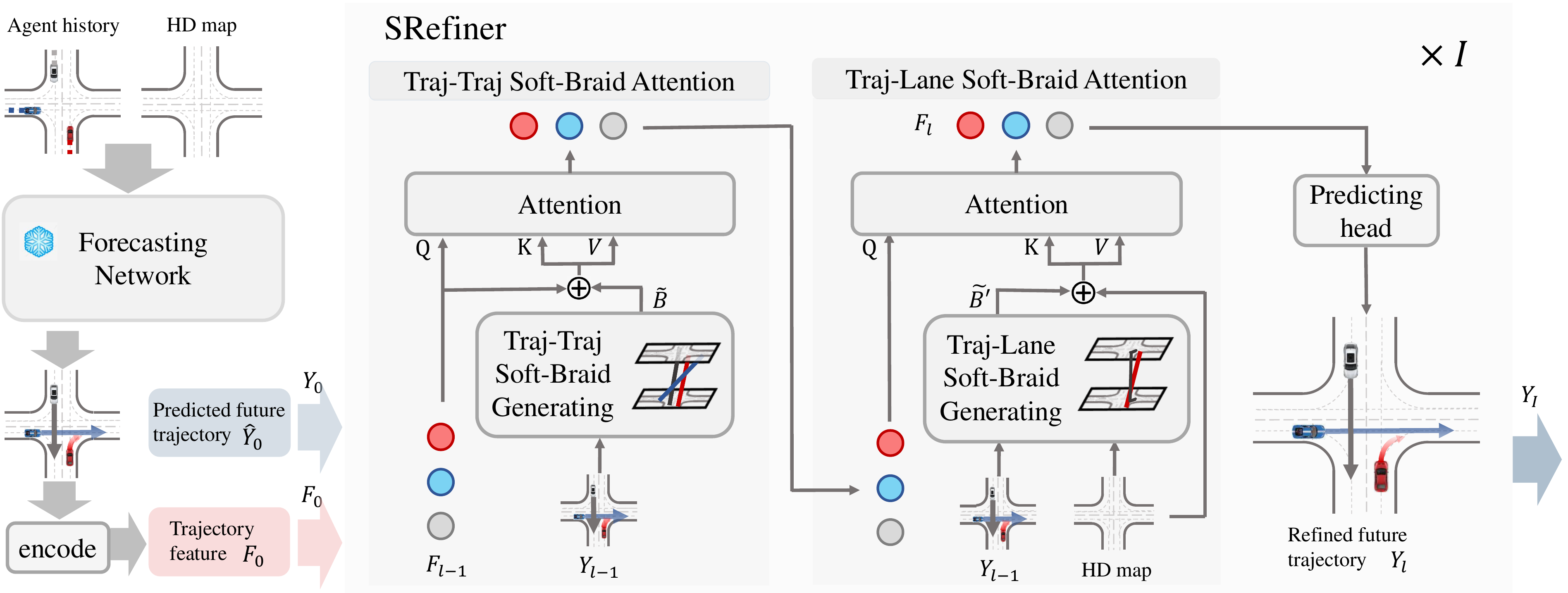}
    \caption{\textbf{Overall pipeline of Soft-Braid Refiner (SRefiner).} We first input historical trajectory information and HD maps into a motion forecasting baseline model. The predicted trajectories from this model are then encoded as initial inputs for SRefiner. SRefiner comprises a Trajectory-Trajectory Soft-Braid Attention Module, a Trajectory-Lane Soft-Braid Attention Module, and a predicting head. The Trajectory-Trajectory module captures soft-braid topological relationships between trajectories to guide refinement, while the Trajectory-Lane module extends this topology to model trajectory-lane interactions. The predicting head subsequently outputs refined trajectories. SRefiner iteratively performs this refinement process $I$ times ($I=3$), updating the soft-braid topology after each iteration, with the output of the final iteration serving as the refined result.}
    \label{fig:pipeline}
\end{figure*}



\noindent\textbf{Trajectory-trajectory soft-braid attention.}
Let $ F \in R^{K \times N \times D}$ denote the trajectory embeddings of a future scenario. Given the embeddings of the $ i-th $ and $ j-th$ trajectories $ F_i,F_j \in R^D$, we employ multi-head cross-attention (MHCA) to facilitate information interaction among trajectories:
\begin{equation}
    \begin{aligned}
        F_i = \texttt{MHCA}(&Q:F_i, \\
        &K:\{F_j + \varphi(\tilde{\sigma}_{i \leftarrow j})\}_{j \in \Omega(i)}, \\
        &V:\{F_j + \varphi(\tilde{\sigma}_{i \leftarrow j})\}_{j \in \Omega(i)}
        ),
        \\
    \end{aligned}
\end{equation}
where $ \varphi(\cdot) $ is a 3-layer MLP, and $\Omega(i) = \left\{j \left| d_{ij} \le \tau_a \right.\right\} $ represents the set of trajectories within the neighborhood of $y_i$. The threshold $\tau_a$ is set to $50$ by default.

\noindent\textbf{Trajectory-lane soft-braid attention.} 
We extend the soft-braid topology to model the dynamic topological relationship between trajectories and lanes.
Specifically, we define the waypoint on trajectory \(y_i\) that is closest to a lane \(L_k\) as the soft intersection point \(C_{i_k}\), and the time at which this minimum distance occurs as \(t_{ik}\). We have: 
\begin{equation}
    t_{ik} = \mathop{\arg\min}\limits_{t} ||y_i(t) - L_k ||,
\end{equation}
\begin{equation}
    C_{i_k} = y_i(t_{ik}),
\end{equation}
where $ L_k $ represents the global coordinate of lane $ L_k $. We characterize the spatial relationship between \(C_{ik}\) and \(L_{k}\) using the distance $d_{ik}$ and orientation $ \theta_{ik}$ of their connecting line:
\begin{equation}
    d_{ik} = || L_k - C_{i_k}||, \theta_{ik} = \arctan (L_k - C_{i_k}).
\end{equation}
We characterize the soft-braid topology between \(y_i\) and \(L_k\) by consifering the velocity and acceleration of agent \(i\) at the soft interaction point \(C_{i_k}\), along with the spatial relationship between \(C_{i_k}\) and lane \(L_k\):
\begin{equation}
    \tilde{\lambda}_{i \leftarrow k} = [\dot{y}^{(i)}_i(t_{ik}), \ddot{y}^{(i)}_i(t_{ik}), d_{ik}, \theta_{ik}^{(i)}].
\end{equation}
The motion state of agent \(i\) and the spatial relationship between \(C_{i_k}\) and lane \(L_k\) are both characterized in the local coordinate system of agent \(i\). $ \widetilde{B}' = \{ \tilde{\lambda}_{i \leftarrow k} \}_{N \times M}, 1 \le i \le N, 1 \le k \le M$ represents the soft-braid topology among all $N$ trajectories and $ M $ lanes in a future scenario. We employ MHCA to facilitate information interaction among trajectories and the HD map:
\begin{equation}
    \begin{aligned}
        F_i = \texttt{MHCA}(&Q:F_i, \\
        &K:\{\varphi(L_k^{(i)} + \tilde{\lambda}_{i \leftarrow k})\}_{k \in \Omega(i)}, \\
        &V:\{\varphi(L_k^{(i)} + \tilde{\lambda}_{i \leftarrow k})\}_{k \in \Omega(i)}
        ),
        \\
    \end{aligned}
\end{equation}
where $ \Omega(i) = \left\{k \left| d_{ik} \le \tau_l \right.\right\} $ is the set of trajectories within the neighborhood of $y_i$. The threshold $\tau_l$ is set to 10 by default.
\definecolor{lightblue}{RGB}{239,245,251}  
\newcommand{\blue}[1]{\cellcolor{lightblue}{#1}}
\begin{table*}[htbp]
\centering
\scriptsize
\caption{\textbf{Performance comparison on the validation/test set of Argoverse v2~\cite{argoversev2} and INTERACTIONS~\cite{interaction} datasets}. Our method brings significant performance improvements for the four baselines on both datasets.}

\begin{tabular}{@{}llllllll@{}}
\toprule
Dataset & Method  & \multicolumn{3}{c}{Validation Set} & \multicolumn{3}{c}{Test Set} \\ 
\midrule
\multirow{5}{*}{Argoverse v2~\cite{argoversev2}} 
& & avgMinFDE $\downarrow$ & avgMinADE $\downarrow$ & actorMR $\downarrow$ & avgMinFDE $\downarrow$ & avgMinADE $\downarrow$ & actorMR $\downarrow$ \\ 
\cmidrule(lr){3-5} \cmidrule(lr){6-8}
& FJMP~\cite{fjmp}         & 1.920 & 0.819 & 0.235 & 1.890 & 0.810 & 0.230 \\ 
& FJMP w/ SRefiner & \blue{1.736 \textcolor{blue}{(-9.6\%)}} & \blue{0.747 \textcolor{blue}{(-8.8\%)}} & \blue{0.221 \textcolor{blue}{(-6.0\%)}} & \blue{1.719 \textcolor{blue}{(-9.1\%)}} & \blue{0.747 \textcolor{blue}{(-7.8\%)}} & \blue{0.213 \textcolor{blue}{(-7.4\%)}} \\ 
\cmidrule(lr){2-8}
& Forecast-MAE~\cite{forecastmae}         & 1.642 & 0.717 & 0.194 & 1.679 & 0.735 & 0.197 \\ 
& Forecast-MAE w/ SRefiner & \blue{1.477 \textcolor{blue}{(-10.1\%)}} & \blue{0.658 \textcolor{blue}{(-8.3\%)}} & \blue{0.183 \textcolor{blue}{(-5.7\%)}} & \blue{1.521 \textcolor{blue}{(-9.4\%)}} & \blue{0.678 \textcolor{blue}{(-7.8\%)}} & \blue{0.186 \textcolor{blue}{(-5.6\%)}} \\ 
\midrule

\multirow{7}{*}{INTERACTIONS~\cite{interaction}} 
& & minJointFDE $\downarrow$ & minJointADE $\downarrow$ & minJointMR $\downarrow$ & minJointFDE $\downarrow$ & minJointADE $\downarrow$ & minJointMR $\downarrow$ \\
\cmidrule(lr){3-5} \cmidrule(lr){6-8}
& AutoBots~\cite{autobots}         & 0.683 & 0.212 & 0.136 & 1.015 & 0.312 & 0.211 \\ 
& AutoBots w/ SRefiner & \blue{0.611 \textcolor{blue}{(-10.5\%)}} & \blue{0.185 \textcolor{blue}{(-12.6\%)}} & \blue{0.119 \textcolor{blue}{(-12.2\%)}} & \blue{0.906 \textcolor{blue}{(-10.7\%)}} & \blue{0.271 \textcolor{blue}{(-13.2\%)}} & \blue{0.175 \textcolor{blue}{(-17.0\%)}} \\ 
\cmidrule(lr){2-8}
& FJMP~\cite{fjmp}         & 0.630 & 0.190 & 0.122 & 0.945 & 0.283 & 0.186 \\ 
& FJMP w/ SRefiner & \blue{0.579 \textcolor{blue}{(-8.1\%)}} & \blue{0.170 \textcolor{blue}{(-10.5\%)}} & \blue{0.110 \textcolor{blue}{(-9.9\%)}} & \blue{0.867 \textcolor{blue}{(-8.3\%)}} & \blue{0.257 \textcolor{blue}{(-9.2\%)}} & \blue{0.163 \textcolor{blue}{(-12.4\%)}} \\ 
\cmidrule(lr){2-8}
& HPNet~\cite{hpnet}         & 0.558 & 0.174 & 0.108 & 0.823 & 0.255 & 0.159 \\ 
& HPNet w/ SRefiner & \blue{0.548 \textcolor{blue}{(-1.8\%)}} & \blue{0.169 \textcolor{blue}{(-3.0\%)}} & \blue{0.104 \textcolor{blue}{(-3.8\%)}} & \blue{0.797 \textcolor{blue}{(-3.2\%)}} & \blue{0.247 \textcolor{blue}{(-3.2\%)}} & \blue{0.152 \textcolor{blue}{(-4.5\%)}} \\ 

\bottomrule
\end{tabular}
\label{tab:relative}

\end{table*}

\begin{table*}[htbp]
\centering
\small
\addtolength{\tabcolsep}{-4pt}
\caption{\textbf{Performance comparison with previous trajectory refinement methods.} Our method achieves state-of-the-art (SOTA) performance compared to other refinement approaches. We evaluate model latency on an RTX 3090 GPU. The latency below is the average inference time of one scenario. SmartRefine's official code refines one agent's trajectory per forward pass. To achieve multi-agent refinement, we use it to refine all agents' trajectories in the scenario independently. $N$ is the number of agents in a scenario.}
\renewcommand\arraystretch{1.05}
\begin{tabular}{@{}lcccccccc@{}}
\toprule
\multirow{2}{*}{Refine Method}  & \multicolumn{4}{c}{ Baseline: Forecast-MAE~\cite{forecastmae}} & \multicolumn{4}{c}{Baseline: FJMP~\cite{fjmp}} \\ 
                                & \multicolumn{4}{c}{Dataset: Argoverse v2~\cite{argoversev2}}  & \multicolumn{4}{c}{Dataset: INTERACTIONS~\cite{interaction}} \\ 
\midrule
 
& avgMinFDE $\downarrow$ & avgMinADE $\downarrow$ & actorMR $\downarrow$ & latency & minJointFDE $\downarrow$ & minJointADE $\downarrow$ & minJointMR $\downarrow$  & latency \\ 
\cmidrule(lr){2-5} \cmidrule(lr){6-9}
Baseline        & 1.642 & 0.717 & 0.194 & -- & 0.630 & 0.190 & 0.122 & -- \\
DCMS~\cite{dcms}            & 1.601 & 0.702 & 0.190 & 5ms & 0.615 & 0.184 & 0.118  & 3ms \\ 
R-Pred~\cite{rpred}          & 1.554 & 0.683 & 0.187 & 12ms & 0.603 & 0.180 & 0.115 & 8ms \\
QCNet~\cite{qcnet}           & 1.520 & 0.674 & 0.185 & 58ms & 0.588 &  0.176 & 0.113 & 33ms \\
SmartRefine~\cite{smartrefine}     & 1.512 & 0.679 & 0.185 & 24$\times N$ms & 0.592 & 0.175 & 0.113 & 14$\times N$ms \\
MTR++~\cite{mtr++}           & 1.495 & 0.670 & \textbf{0.183} & 54ms & 0.585 & 0.172 & 0.111 & 30ms \\
SRefiner (Ours) & \textbf{1.477} & \textbf{0.658} & \textbf{0.183} & 28ms & \textbf{0.579} & \textbf{0.170} & \textbf{0.110} & 18ms \\

\bottomrule
\end{tabular}
\label{tab:compare}

\end{table*}

\subsection{Soft-Braid Refiner}

We propose a $I$-iteration trajectory refiner named Soft-Braid Refiner (SRefiner). SRefiner takes the future trajectories predicted by a baseline network as input and employs soft-braid attention to model the interactions among trajectories as well as between trajectories and the map, resulting in more accurate future trajectories. Specifically, we first encode initial trajectories $ Y_0$ as $ F_{0}$:
\begin{equation}
    F_{0} = \varphi([O, \theta, \texttt{PE}(Y_0)]).
\end{equation}
Instead of using the soft-braid topology derived from the initial trajectories $ Y_0$ to guide all iterations, we adopt a progressive strategy. That is, we use the soft-braid topology from the trajectories refined in the $(l-1)$-th iteration to guide those in the $l$-th iteration. This strategy allows for iterative refinement based on progressively improved topological information, thereby enhancing the accuracy and relevance of each subsequent iteration. The progressive refinement process as follows:
\begin{equation}
    \widetilde{B}_{l-1}, \widetilde{B}'_{l-1} = \mathcal{S}(Y_{l-1}, L),
\end{equation}
\begin{equation}
    F_{l} = \texttt{SoftBraidAttn}(F_{l-1}, L, \widetilde{B}_{l-1}, \widetilde{B}'_{l-1}),
\end{equation}
\begin{equation}
    Y_l = \varphi(F_l) + Y_{l-1},
\end{equation}
where \(\mathcal{S}(\cdot)\) denotes the computation process of the soft-birad topology, \(\widetilde{B}_{l-1}\) and \(\widetilde{B}'_{l-1}\) represent the soft-braid topology between the trajectories and between the trajectories and the map lanes from the $(l-1)$-th iteration, respectively. The $\texttt{SoftBraidAttn}(\cdot)$ comprises sequential trajectory-trajectory soft-braid attention and trajectory-lane soft-braid attention, as illustrated in Figure~\ref{fig:pipeline}. An MLP then predicts the refined output for the $l$-th iteration. This procedure is iterated $I$ times to achieve the final refinement.

\subsection{Optimization}
We supervise the outputs of all $ I $ iterations $ \{Y_l\}_{l=1}^I$. Given that our method focuses on joint trajectory prediction for multiple agents, we adhere to the joint prediction paradigm and employ a joint winner-takes-all (WTA) loss. In joint prediction, we consider the predictions of all agents within the same mode as a single predicted world. Therefore, the $k_l$-th mode to be optimized is determined by minimizing the joint average displacements between the predicted world and the ground truth.
Specifically, for the $l$-th iteration output \(Y_l\), the optimal mode \(k_l\) is determined as:
\begin{equation}
    k_l =\mathop{\arg\min}\limits_{k \in [1,K]} \frac{1}{N} \sum_{i=1}^{N} \| Y_{l,i,k} - Y_{\text{gt},i} \|, 
\end{equation}
where \(Y_{l,i,k}\) is the predicted trajectory of agent \(i\) in mode \(k\) at the $l-th$ iteration, and \(Y_{i, \text{gt}}\) is the ground truth trajectory of agent \(i\). We employ Huber Loss to supervise \(Y_l\):
\begin{equation}
    \mathcal{L}_{l} = \frac{1}{N} \sum_{i=1}^{N} \mathcal{L}_{Huber}(Y_{l,i,k_l} - Y_{i, \text{gt}}).
\end{equation}
Overall, the final loss function of the entire model is formulated as follows:
\begin{equation}
    \mathcal{L} = \frac{1}{I} \sum_{l=1}^{I} \mathcal{L}_{l}.
\end{equation}

\section{Experiments}

\begin{figure*}[htbp]
    \centering
    \includegraphics[width=\linewidth]{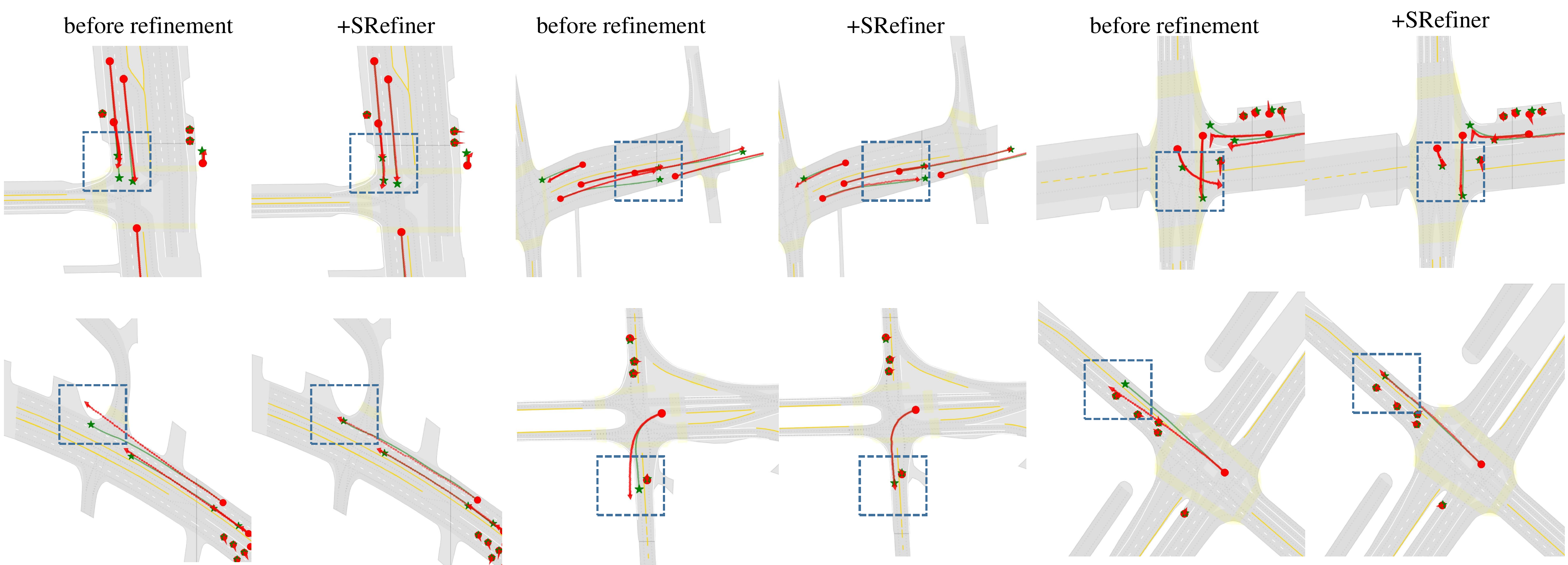}
    \caption{\textbf{Visualization of refinement results. }Red lines indicate predicted future trajectories, with red dots marking their starting points, while green lines and pentagrams represent ground truth (GT) future trajectories and endpoints. The first row demonstrates that SRefiner effectively reduces trajectory collisions and refines safer driving behaviors. The second row illustrates its capability to correct trajectories that deviate from drivable areas or intrude into opposing lanes.}
    \label{fig:vis}
\end{figure*}

\begin{table*}[ht]
\centering
\small
\caption{\textbf{Effect of each components of SRefiner.} The topology update strategy involves using the refined trajectory topology from the previous iteration to guide the current iteration's refinement.}
\begin{tabular}{@{}ccccccc@{}}
\toprule

Method & Traj-Traj Soft-Braid & Traj-Lane Soft-Braid & Topology Update & avgMinFDE $\downarrow$ & avgMinADE $\downarrow$ & actorMR $\downarrow$ \\
\midrule
 
M1 &            &            &            & 1.642 & 0.717 & 0.194  \\
M2 &            & \checkmark & \checkmark & 1.514 & 0.678 & 0.184  \\
M3 & \checkmark &            & \checkmark & 1.497 & 0.670 & \textbf{0.183}  \\ 
M4 & \checkmark & \checkmark &            & 1.522 & 0.673 & 0.186  \\
M5 & \checkmark & \checkmark & \checkmark & \textbf{1.477} & \textbf{0.658} & \textbf{0.183}  \\

\bottomrule
\end{tabular}
\label{tab:component}

\end{table*}

\subsection{Experimental Settings}

\textbf{Dataset.} We evaluate the proposed method on two widely-used multi-agent motion forecasting datasets: \textit{Argoverse v2}~\cite{argoversev2} and \textit{INTERACTION}~\cite{interaction}, both of which feature rich vehicle interaction scenarios and are sampled at 10 Hz. For \textit{Argoverse v2}, the task focuses on long-term motion forecasting, predicting 6-second future trajectories based on 5 seconds of historical observations (\(T_-\) = 50, \(T_+\) = 60). As for \textit{INTERACTION}, it emphasizes short-term forecasting, predicting 3 second future trajectories from 1 second of history (\(T_-\) = 10, \(T_+\) = 30). Both datasets provide high-definition (HD) maps.

\noindent\textbf{Metrics.} For evaluation, we adopt official multi-world trajectory forecasting metrics. Specifically for Argoverse v2, we use the Average Minimum Final Displacement Error (\textbf{avgMinFDE}), Average Minimum Average Displacement Error (\textbf{avgMinADE}), and Actor Miss Rate (\textbf{actorMR}). \textbf{avgMinFDE} represents the mean final displacement error across all scored actors in the predicted world with the lowest overall FDE, considered the ``best" world. \textbf{avgMinADE} measures the mean average displacement error. \textbf{actorMR} calculates the proportion of scored actors with a final displacement error exceeding $2$ meters in the ``best" world. For the INTERACTION dataset, we apply \textbf{minJointADE}, \textbf{minJointFDE}, and \textbf{minJointMR} to evaluate joint trajectory predictions. These metrics are analogous to those in Argoverse v2 but differ in that they assess all actors rather than only ``scored" ones. For more details, please refer to the supplementary materials.

\noindent\textbf{Baselines.} Our SRefiner can be seamlessly integrated into most existing trajectory prediction methods to boost multi-agent trajectory forecasting accuracy. In our experiment, we evaluate the performance improvements provided by SRefiner using four open-source and widely-used baselines: AutoBots~\cite{autobots}, FJMP~\cite{fjmp}, Forecast-MAE~\cite{forecastmae}, HPNet~\cite{hpnet}.


\subsection{Main Results}

\subsubsection{Improvement Over Baselines}
We report the relative improvements of SRefiner over four baselines across the validation and test sets of two datasets, as shown in Table.~\ref{tab:relative}. It is evident that our SRefiner provides significant and consistent gains. For instance, on the Argoverse v2 test set, SRefiner reduces the avgMinFDE by $ 9.1\% $ for FJMP and by $ 9.4\% $ for Forecast-MAE. On the INTERACTIONS test set, it decreases the minJointFDE by $ 10.7\% $ for AutoBot, $ 8.3\% $ for FJMP, and $ 3.2\% $ for HPNet. Notably, HPNet combined with SRefiner achieves state-of-the-art (SOTA) performance on the INTERACTIONS dataset.

\subsubsection{Comparison with Refinement Methods}
We compare SRefiner with previous trajectory refinement methods on the validation sets of two datasets, as shown in Table~\ref{tab:compare}. Since DCMS, R-Pred, and MTR++ are not open source, we reproduce these methods within our framework following their described approaches to ensure a fair and convenient comparison. SmartRefine's official code\footnote{https://github.com/opendilab/SmartRefine} supports only single-agent refinement, so we process each vehicle in the scene individually to obtain multi-agent refinement results. In contrast, our SRefiner can refine the trajectories of all vehicles in the scene simultaneously. As illustrated in Table~\ref{tab:compare}, SRefiner achieves state-of-the-art (SOTA) performance compared to prior refinement methods.

\subsubsection{Qualitative Results}
As shown in Row $1$ of Fig.~\ref{fig:vis}, SRefiner enhances inter-trajectory awareness, effectively reducing collisions and refining safer future trajectories. Row $2$ demonstrates its capability to correct trajectories that deviate from drivable areas or encroach into opposing lanes through trajectory-map interactions, thereby generating safer trajectories.

\subsection{Ablation Study}

\begin{table}[t]
\centering
\small
\renewcommand\arraystretch{1.2}
\caption{Ablation study on the effect of soft-braid topology. $\dagger$ indicates only model the topology between trajectories.}
\begin{tabular}{@{}lc@{}}
\toprule

Method & avgMinFDE $\downarrow$ \\
\midrule
Baseline & 1.642 \\
No Topology & 1.530  \\
Braid Topology (BeTop~\cite{betop}) & 1.512  \\ 
Soft-Braid Topology (Ours)$\dagger$ & 1.497 \\
Soft-Braid Topology (Ours) & \textbf{1.477}  \\

\bottomrule
\end{tabular}
\label{tab:topology}

\end{table}

In this section, we validate the efficacy of each component of SRefiner, investigate the impact of soft-braid topology, and analyze hyperparameter selection. Experiments are conducted using Forecast-MAE as the baseline on the Argoverse v2 validation set. Additional ablation studies are provided in the supplementary materials.

\noindent\textbf{Effect of each components of SRefiner.} We evaluate the effectiveness of each component of SRefiner, including Trajectory-Trajectory Soft-Braid Attention, Trajectory-Lane Soft-Braid Attention, and the topology update strategy. The topology update strategy involves using the refined trajectory topology from the previous iteration to guide the current layer's refinement, rather than relying on the initially proposed trajectory topology for all iterations. Table~\ref{tab:component} demonstrates that each component significantly contributes to the overall performance. 

\noindent\textbf{Effect of soft-braid topology.} 
Braid topology~\cite{braid1947theory} models the intersection relationship between two future trajectories: intersection and non-intersection. BeTop~\cite{betop} directly applies the braid topology to enable information fusing among intersecting trajectories and disable those among non-intersecting ones, which is termed braid attention. Such braid attention leading to more accurate refinement compared to implicit interaction modeling without any guidance of topological information. Furthermore, our soft-braid topology models the spatio-temporal topological features among all trajectories, resulting in more significant performance improvements, as shown in Table~\ref{tab:topology}.

\begin{figure}[t]
    \centering
    \includegraphics[width=0.98\linewidth]{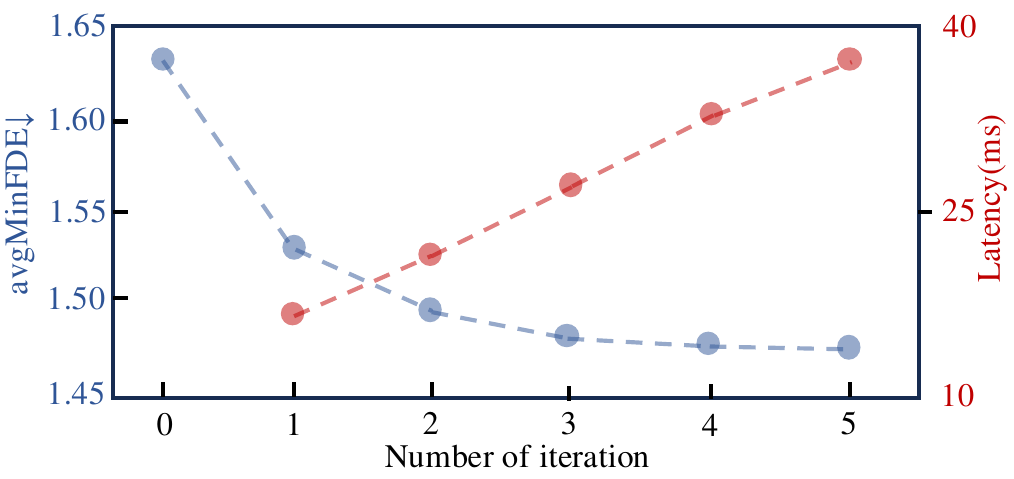}
    \caption{Ablation study of the number of iteration. To balance performance and latency, we set the iteration number to $3$.}
    \label{fig:iteration}
\end{figure}

\noindent\textbf{Ablation study on hyperparameters.} We vary the local radius of Trajectory-Trajectory Soft-Braid Attention and Trajectory-Lane Soft-Braid Attention. We find that the performance of our method is not sensitive to the selection of them. Consequently, we set them to $50$m and $10$m by default, as shown in Table~\ref{table:radius}. Regarding the number of refinement iterations, we observe that setting it to $3$ yields optimal results, as increasing the number of iterations beyond $3$ does not result in significant performance improvements but incurs additional computational costs, as shown in Figure~\ref{fig:iteration}.

\begin{table}[t]
    \footnotesize
    \caption{Ablation study on the Trajectory-Trajectory Soft-Braid Attention local radius and Trajectory-Lane Soft-Braid Attention local radius. ``T-T local radius" and ``T-L local radius" refer to $ \tau_a $ and $ \tau_l $ in Sec.~\ref{Soft Braid Attention}}
    \label{table:radius}
    \begin{minipage}[t]{0.23\textwidth}
        \renewcommand\arraystretch{1.2}
        \vspace{0pt}
        \centering
        
        \label{radius-tt}
        \begin{tabular}{@{}cc@{}}
        \toprule
        T-T local radius           &    avgMinFDE $\downarrow$ \\
        \midrule
        10               &    1.489                \\
        30      &    1.482                 \\
        50      &    \textbf{1.477}                \\
        100      &    1.480                \\
        \bottomrule
        \end{tabular}

    \end{minipage}%
    \begin{minipage}[t]{0.23\textwidth}
        \renewcommand\arraystretch{1.2}
        \vspace{0pt}
        \centering
        
        \label{radius-tl}
        \begin{tabular}{@{}cc@{}}
        \toprule
        T-L local radius           &    avgMinFDE $\downarrow$ \\
        \midrule
        2               &    1.484                \\
        5      &    1.482                 \\
        10      &    \textbf{1.477}                \\
        20      &    \textbf{1.477}                \\
        \bottomrule
        \end{tabular}
    \end{minipage}

\end{table}




\section{Conclusion}


We present Soft-Braid Refiner (SRefiner), a topology-aware framework for trajectory refinement that addresses the challenge of modeling future spatio-temporal interactions. SRefiner introduces Trajectory-Trajectory Soft-Braid Attention, dynamically capturing trajectory-topology relationships via soft intersection points, and extends this to Trajectory-Lane Soft-Braid Attention for modeling trajectory-lane interactions. By iteratively refining trajectories while updating topological information, SRefiner achieves SOTA performance in trajectory refinement methods. This work highlights the importance of explicit topology modeling in trajectory prediction and lays the groundwork for applying braid-inspired structures to autonomous systems.

{
    \small
    \bibliographystyle{ieeenat_fullname}
    \bibliography{main}
}
\clearpage
\setcounter{page}{1}
\maketitlesupplementary


The overall structure of the supplementary material is listed as follows:

$\triangleright$ Sec.~\ref{sec:sup_metrics}: \textit{Details of evaluation metrics.}

$\triangleright$ Sec.~\ref{sec:sup_discussion}: \textit{Discussion with BeTop~\cite{betop}.}

$\triangleright$ Sec.~\ref{sec:sup_inplementation}: \textit{Implementation details.}

$\triangleright$ Sec.~\ref{sec:sup_ablation}: \textit{Ablation study on other baseline and dataset.}

\begin{table*}[!t]
\centering
\small
 \renewcommand{\dblfloatpagefraction}{.9}
\caption{Ablation study of the effect of each components of SRefiner with FJMP~\cite{fjmp} on Argoverse v2~\cite{argoversev2}.}
\begin{tabular}{@{}ccccccc@{}}
\toprule
Method & Traj-Traj Soft-Braid & Traj-Lane Soft-Braid & Topology Update & avgMinFDE $\downarrow$ & avgMinADE $\downarrow$ & actorMR $\downarrow$ \\
\midrule
 
M1 &            &            &            & 1.920 & 0.819 & 0.235  \\
M2 &            & \checkmark & \checkmark & 1.780 & 0.772 & 0.223  \\
M3 & \checkmark &            & \checkmark & 1.756 & 0.759 & 0.222  \\
M4 & \checkmark & \checkmark &            & 1.787 & 0.765 & 0.226  \\
M5 & \checkmark & \checkmark & \checkmark & \textbf{1.736} & \textbf{0.747} & \textbf{0.221}  \\
\bottomrule
\end{tabular}
\label{tab:sup_component_av2}
\end{table*}

\begin{table*}[!t]
\centering
\small
 \renewcommand{\dblfloatpagefraction}{.9}
\caption{Ablation study of the effect of each components of SRefiner with FJMP~\cite{fjmp} on INTERACTIONS~\cite{interaction}.}
\begin{tabular}{@{}ccccccc@{}}
\toprule

Method & Traj-Traj Soft-Braid & Traj-Lane Soft-Braid & Topology Update & avgMinFDE $\downarrow$ & avgMinADE $\downarrow$ & actorMR $\downarrow$ \\
\midrule
 
M1 &            &            &            & 0.630 & 0.190 & 0.122  \\
M2 &            & \checkmark & \checkmark & 0.591 & 0.178 & 0.112  \\
M3 & \checkmark &            & \checkmark & 0.585 & 0.174 & \textbf{0.110}  \\ 
M4 & \checkmark & \checkmark &            & 0.593 & 0.175 & 0.114  \\
M5 & \checkmark & \checkmark & \checkmark & \textbf{0.579} & \textbf{0.170} & \textbf{0.110}  \\
\bottomrule
\end{tabular}
\label{tab:sup_component_interactions}
\end{table*}

\begin{table*}[!t]
    \small
     \renewcommand{\dblfloatpagefraction}{.9}
    \begin{minipage}{0.45\textwidth}
        \renewcommand\arraystretch{1.2}
        \vspace{0pt}
        \centering
        \caption{Ablation study on the effect of soft-braid topology with FJMP~\cite{fjmp} on Argoverse v2~\cite{argoversev2}. $ \dagger $ indicates only model the topology between trajectories.}
        \label{table:sup_braid_av2}
        \begin{tabular}{@{}lc@{}}
        \toprule
        Method & avgMinFDE $\downarrow$ \\
        \midrule
        Baseline & 1.920 \\
        No Topology & 1.794  \\
        Braid Topology (BeTop~\cite{betop}) & 1.770  \\ 
        Soft-Braid Topology (Ours)$ \dagger $ & 1.756  \\
        Soft-Braid Topology (Ours) & \textbf{1.736}  \\
\bottomrule
\end{tabular}
    \end{minipage}%
    \hspace{0.5cm}
    \begin{minipage}{0.45\textwidth}
        \renewcommand\arraystretch{1.2}
        \vspace{0pt}
        \centering
        \caption{Ablation study on the effect of soft-braid topology with FJMP~\cite{fjmp} on INTERACTIONS~\cite{interaction}. $ \dagger $ indicates only model the topology between trajectories.}
        \label{table:sup_braid_interactions}
        \begin{tabular}{@{}lc@{}}
        \toprule
        Method & minJointFDE $\downarrow$ \\
        \midrule
        Baseline & 0.630 \\
        No Topology & 0.601  \\
        Braid Topology (BeTop~\cite{betop}) & 0.594  \\ 
        Soft-Braid Topology (Ours)$ \dagger $ & 0.585  \\
        Soft-Braid Topology (Ours) & \textbf{0.579}  \\
        \bottomrule
        \end{tabular}
    \end{minipage}
\end{table*}

\begin{table*}[!t]
    \footnotesize
     \renewcommand{\dblfloatpagefraction}{.9}
    \begin{minipage}[t]{0.46\textwidth}
        \caption{Ablation study on the Trajectory-Trajectory Soft-Braid Attention local radius and Trajectory-Lane Soft-Braid Attention local radius with FJMP~\cite{fjmp} on Argoverse v2~\cite{argoversev2}. 
        }
        \label{table:sup_radius_av2}
        \begin{minipage}{0.22\textwidth}
            \renewcommand\arraystretch{1.2}
            \vspace{0pt}
            \centering
            \setlength{\tabcolsep}{1mm}{
            \begin{tabular}{@{}cc@{}}
            \toprule
            T-T local radius           &    avgMinFDE $\downarrow$ \\
            \midrule
            10      &    1.752                \\
            30      &    1.739                 \\
            50      &    \textbf{1.736}                \\
            100     &    \textbf{1.736}                \\
            \bottomrule
            \end{tabular} }
        \end{minipage}%
        \hspace{2cm}
        \begin{minipage}{0.22\textwidth}
            \renewcommand\arraystretch{1.2}
            \vspace{0pt}
            \centering
            \setlength{\tabcolsep}{1mm}{
            \begin{tabular}{@{}cc@{}}
            \toprule
            T-L local radius           &    avgMinFDE $\downarrow$ \\
            \midrule
            2               &    1.746                \\
            5      &    1.738                 \\
            10      &    \textbf{1.736}                \\
            20      &    1.738                \\
            \bottomrule
            \end{tabular}}
        \end{minipage}
    
    \end{minipage}%
    \hspace{0.2cm}
    \begin{minipage}[t]{0.48\textwidth}
        \caption{Ablation study on the Trajectory-Trajectory Soft-Braid Attention local radius and Trajectory-Lane Soft-Braid Attention local radius with FJMP~\cite{fjmp} on INTERACTIONS~\cite{interaction}. 
        }
        \label{table:sup_radius_interactions}
        \begin{minipage}{0.23\textwidth}
            \renewcommand\arraystretch{1.2}
            \vspace{0pt}
            \centering
            \setlength{\tabcolsep}{1mm}{
            \begin{tabular}{@{}cc@{}}
            \toprule
            T-T local radius           &    minJointFDE $\downarrow$ \\
            \midrule
            10      &    0.585                \\
            30      &    0.582                 \\
            50      &    \textbf{0.579}                \\
            100     &    \textbf{0.579}                \\
            \bottomrule
            \end{tabular} }
        \end{minipage}%
        \hspace{2.1cm}
        \begin{minipage}{0.23\textwidth}
            \renewcommand\arraystretch{1.2}
            \vspace{0pt}
            \centering
            \setlength{\tabcolsep}{1mm}{
            \begin{tabular}{@{}cc@{}}
            \toprule
            T-L local radius           &    minJointFDE $\downarrow$ \\
            \midrule
            2               &    0.584                \\
            5      &    0.580                 \\
            10      &    \textbf{0.579}                \\
            20      &    0.580                \\
            \bottomrule
            \end{tabular}}
    \end{minipage}
    \end{minipage}
\end{table*}

\section{Evaluation metrics}
\label{sec:sup_metrics}

\noindent We first introduce three evaluation metrics used in the Argoverse v2 dataset~\cite{argoversev2}: avgMinFDE, avgMinADE and actorMR.

\noindent\textbf{avgMinFDE.} The average Final Displacement Error (FDE) associated with the predicted world that has the lowest mean FDE among all ( K ) predicted worlds. FDE is defined as the L2 distance between the endpoint of the predicted trajectory and the ground truth. The mean FDE is the FDE averaged across all scored actors within a scenario. The index of the world with the lowest mean FDE is determined as follows: 

\begin{equation}
    idx_{FDE} =\mathop{\arg\min}\limits_{k \in [1,K]} \frac{1}{N} \sum_{i=1}^{N} \| Y_{i,k}(T_+) - Y_{i, \text{gt}}(T_+) \|, 
\end{equation}
where $ Y_{i,k}(T_+) $ is the endpoint of the $i-th$ predicted trajectory in the $k-th$ mode, and $ Y_{i, \text{gt}}(T_+) $ is the endpoint of the $i-th$ ground truth trajectory. The avgMinFDE is then calculated as:
\begin{equation}
    \text{avgMinFDE} = \frac{1}{N} \sum_{i=1}^{N} \|Y_{i,idx_{FDE}}(T_+) - Y_{i,\text{gt}}(T_+)\|.
\end{equation}

\noindent\textbf{avgMinADE.}The mean ADE(Average Displacement Error) is associated with a predicted world that has the lowest mean ADE among all $K$ predicted worlds. ADE is defined as the mean L2 distance between the predicted trajectory and the ground truth. The mean ADE is the ADE averaged across all scored actors within a scenario. The index of the world with the lowest mean ADE is determined as follows:

\begin{equation}
    idx_{ADE} =\mathop{\arg\min}\limits_{k \in [1,K]} \frac{1}{N} \sum_{i=1}^{N} \| Y_{i,k} - Y_{i, \text{gt}} \|, 
\end{equation}
where $ Y_{i,k} $ is the $i-th$ predicted trajectory in the $k-th$ mode, and $ Y_{i, \text{gt}}(T_+) $ is the $i-th$ ground truth trajectory. The avgMinADE is then calculated as:

\begin{equation}
    \text{avgMinFDE} = \frac{1}{N} \sum_{i=1}^{N} \|Y_{i,idx_{ADE}} - Y_{i,\text{gt}}\|.
\end{equation}

\noindent\textbf{actorMR.}The actor Miss Rate (actorMR) is defined as the proportion of actor predictions that are considered to have ``missed''($>2$m FDE) in the the ``best'' (lowest minFDE) predicted world:

\begin{equation}
    \text{actorMR} = \frac{1}{N} \sum_{i=1}^{N} \mathbb{I}_{\{\|Y_{i,idx_{FDE}} - Y_{i,\text{gt}}\| > 2\}}.
\end{equation}

For the INTERACTIONS dataset~\cite{interaction}, the evaluation metric minJointFDE is defined identically to avgMinFDE in the Argoverse v2 dataset. The same applies to minJointADE and avgMinADE. However, the definition of minJointMR in the INTERACTIONS dataset differs from actorMR in the choice of the threshold:

\begin{equation}
    \text{minJointMR} = \frac{1}{N} \sum_{i=1}^{N} \mathbb{I}_{\{\|Y_{i,idx_{FDE}} - Y_{i,\text{gt}}\| > \tau\}},
\end{equation}
where $ \tau $ denotes the miss rate threshold, defined as follows:

\begin{equation}
    \tau = 
    \begin{cases}
    1, &  v  \leq 1.4m/s  \\
    1+\frac{v-1.4}{11-1.4}, & 1.4m/s < v \leq 11m/s \\
    2, & otherwise,
    \end{cases}
\end{equation}
where $ v $ is the ground-truth velocity at the final timestep.

\section{Discussion with BeTop}
\label{sec:sup_discussion}

BeTop~\cite{betop} focuses on the integration of trajectory prediction and planning (IPP) in autonomous driving. It predicts the intersection relationships between future trajectories and uses ground truth (GT) to supervise these relationships. It is a direct application of braid topology. Our SRefiner focuses on multi-agent trajectory refinement. We propose a soft-braid topology to capture the spatio-temporal topological relationships among all predicted future trajectories. Additionally, SRefiner uses the soft-braid topology to model the spatio-temporal topological relationships between future trajectories and lanes, whereas BeTop only models the intersections between trajectories. As demonstrated in Table~\ref{tab:topology}, for the task of trajectory refinement, the performance of our proposed soft-braid attention surpasses that of the braid attention used in BeTop.

\section{Inplementation details}
\label{sec:sup_inplementation}
For both Argoverse v2~\cite{argoversev2} and INTERACTIONS~\cite{interaction} dataset, we train the model for $64$ epochs with a batch size of $16$ on a single RTX $3090$ GPU. We utilize the AdamW optimizer with a cosine learning rate schedule and a weight decay of $0.0001$. For Argoverse v2, the initial learning rate is set to $1 \times 10^{-4}$, while for INTERACTIONS, it is set to $3 \times 10^{-4}$. The embedding dimension of the model is $64$.

\section{More ablation study}
\label{sec:sup_ablation}

In the main paper, we report ablation studies using Forecast-MAE~\cite{forecastmae} on Argoverse v2 dataset~\cite{argoversev2} due to the page limit. Here, we present ablation studies using FJMP~\cite{fjmp} on both two datasets. Tables~\ref{tab:sup_component_av2} and ~\ref{tab:sup_component_interactions} illustrate the ablation study of the impact of each component of SRefiner with FJMP, demonstrating that each component significantly contributes to the overall performance. Tables ~\ref{table:sup_braid_av2} and ~\ref{table:sup_braid_interactions} show the ablation study on the effect of the soft-braid topology with FJMP on the two datasets. The results indicate that using the soft-braid topology to guide the trajectory refinement yields significant performance improvements and outperform the direct use of the braid topology. Tables~\ref{table:sup_radius_av2} and~\ref{table:sup_radius_interactions} present the ablation study on the choice of trajectory-trajectory soft-braid attention local radius $\tau_a$ and trajectory-lane soft-braid attention local radius $\tau_l$, which are set to 50m and 10m by default.

\end{document}